\documentclass[10pt,twocolumn,letterpaper]{article}

\usepackage{cvpr}
\usepackage{times}
\usepackage{epsfig}
\usepackage{graphicx}
\usepackage{amsmath}
\usepackage{amssymb}

\usepackage{cite}
\usepackage{stmaryrd}
\usepackage{bbold}
\usepackage[export]{adjustbox}
\usepackage{enumerate}
\usepackage{epstopdf}
\usepackage[normalem]{ulem}
\usepackage{multirow}
\usepackage[percent]{overpic}
\usepackage{algorithm}
\usepackage[noend]{algpseudocode}
\usepackage{comment}

\usepackage[normalem]{ulem} % stricktrough 
\usepackage{color} 								% use color for text

\usepackage[pagebackref=true,breaklinks=true,letterpaper=true,colorlinks,bookmarks=false]{hyperref}

\cvprfinalcopy % *** Uncomment this line for the final submission

\def\cvprPaperID{1175} % *** Enter the CVPR Paper ID here

\ifcvprfinal\fi
\begin{document}

%%%%%%%%% TITLE
\title{Co-Occurrence Filter}

\author{Roy Jevnisek\\
Tel-Aviv University \\ 
{\tt\small jernisek@post.tau.ac.il}
\and
Shai Avidan\\
Tel-Aviv University \\ 
{\tt\small avidan@eng.tau.ac.il}
}

\maketitle
%\thispagestyle{empty}

%%%%%%%%%%%%%%%%%%%%%%%%%%%%%%%%%%%%%%%
% Abstruct
%%%%%%%%%%%%%%%%%%%%%%%%%%%%%%%%%%%%%%%
\begin{abstract}

Co-occurrence Filter (CoF) is a boundary preserving filter. It is based on the Bilateral Filter (BF) but instead of using a Gaussian on the range values to preserve edges it relies on a co-occurrence matrix. Pixel values that co-occur frequently in the image (i.e., inside textured regions) will have a high weight in the co-occurrence matrix. This, in turn, means that such pixel pairs will be averaged and hence smoothed, regardless of their intensity differences. On the other hand, pixel values that rarely co-occur (i.e., across texture boundaries) will have a low weight in the co-occurrence matrix. As a result, they will not be averaged and the boundary between them will be preserved. The CoF therefore extends the BF to deal with boundaries, not just edges. It learns co-occurrences directly from the image. We can achieve various filtering results by directing it to learn the co-occurrence matrix from a part of the image, or a different image. We give the definition of the filter, discuss how to use it with color images and show several use cases.

\end{abstract}
%%%%%%%%%%%%%%%%%%%%%%%%%%%%%%%%%%%%%%%

%%%%%%%%%%%%%%%%%%%%%%%%%%%%%%%%%%%%%%%
% Introduction
%%%%%%%%%%%%%%%%%%%%%%%%%%%%%%%%%%%%%%%
\section{Introduction}

There is a long and rich history of edge-preserving filters. These filters smooth the image while preserving its edges. This begs the question: what is an edge? The overwhelming answer in edge-preserving filter literature is that an edge is a sharp discontinuity in intensity value. 

Recent edge detectors give a different answer to this question. Instead of focusing on edge detection, they focus on boundary detection where the goal is to detect boundaries between textures. That is, edges within texture should be ignored while edges that serve as boundaries between textures should be marked.

Our goal is to design a boundary preserving filter that will smooth edges within a textured region and not across texture boundaries.

Co-occurrence Filter (CoF) is the happy marriage of boundary detection and edge preserving filters. It combines ideas from the edge detection literature directly into the filtering process. As a result, there is no need for a two-stage solution.

We start with the Bilateral Filter (BF), which is a well known edge-preserving filter. The output of the BF at a reference pixel is a weighted average of pixels in its neighborhood. The BF mixes pixel values based on two Gaussians. The spatial Gaussian assigns weight based on proximity in the image plane and the range Gaussian assigns weight based on similarity in appearance. As a result, nearby pixels with small intensity differences will mix, while pixels that are far away or with large intensity difference will not. This gives the BF its edge-preserving power.

Because of the way it is defined, the BF can not distinguish between edges within a texture and edges between textures. This is where Co-occurrence information steps in. The Co-occurrence Filter (CoF), that we propose, replaces the range Gaussian filter of the BF with a normalized co-occurrence matrix. Pixel values that co-occur frequently (i.e., in a textured region) will have a high weight and will therefore mix together. This way texture will be smoothed. On the other hand, pixel values that rarely co-occur (i.e., on the boundary between textures) will have a low weight and will therefore not mix. This way smoothing will not occur across texture boundaries.

\begin{figure*}[t]
\begin{center}
\begin{tabular}{c}
\includegraphics[width=0.97\linewidth]{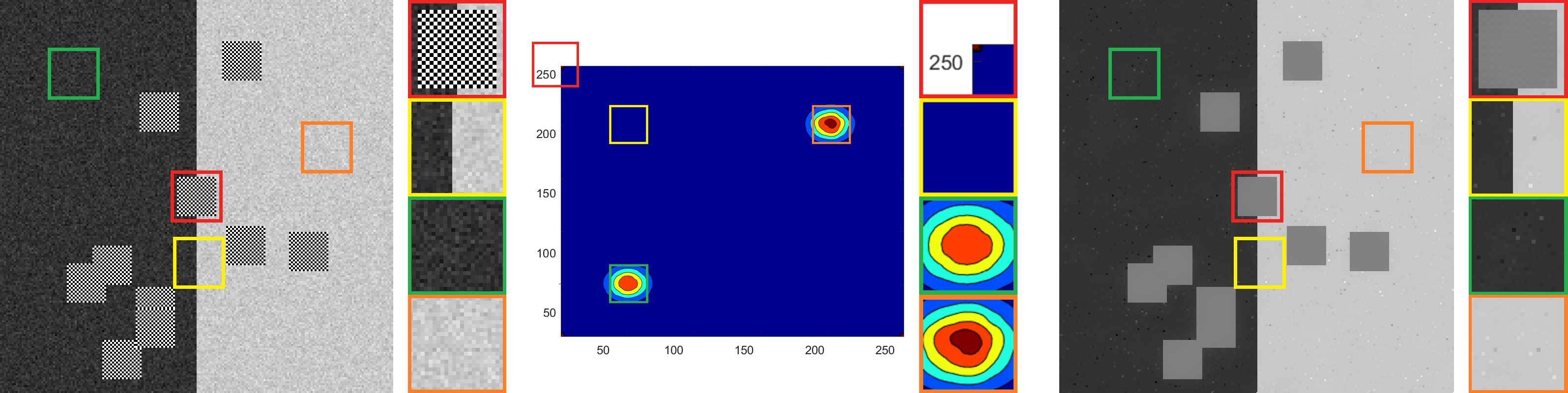}\\

\end{tabular}
\end{center}
\caption{ \textbf{ CoF is not about edge strength:} (Left) input image with zoom-ins. (Center) Co-occurrence matrix with zoom-ins. The two large Gaussian "`bumps"' correspond to the two regions in the image. The size of the Gaussian correlates with the amount of noise added to the image. There is a red dot (i.e., high weight) at the four corners of the co-occurrence matrix - this captures the checkerboard black-white co-occurrences. (Right) CoF result with zoom-ins. The Gaussian noise is removed, the checkerboards are smoothed and the sharp edge between the two regions is preserved.}
\label{fig.teaser_2}
\end{figure*}

Figure~\ref{fig.teaser_2} shows that CoF is about texture and {\em not} about edge strength. The input image consists of two regions (dark on the left side and light on the right side) corrupted by white Gaussian noise. In addition, there are several patches with checkerboard pattern spread across the image plane. The intensity difference between the two regions is lower than that of the checkerboard. The co-occurrence matrix computed from that image gives high weight to the Gaussian noise and the checkerboard texture because they are prevalent in the image. It gives a low weight to the boundary between the two regions of the image, because it is quite a rare phenomenon. CoF filters out the noise and smooths out the checkerboard patches while keeping  sharp boundaries between the different textures.

The proposed filter enjoys a couple of advantages. First, there is no parameter tweaking, as it collects co-occurrence information directly from the image. Second, the user can specify from where the filter should collect the co-occurrence data. For example, the filter can collect data from the whole image, part of it or from a different image altogether.

Extending co-occurrence matrices to deal with color images is not trivial because the co-occurrence space becomes prohibitively large. Simply quantizing RGB values introduces strong aliasing artifacts and we develop an approximation scheme that lets us handle color images gracefully. The resulting filter is fast in practice and can be used in different scenarios and for various artistic effects.

%%%%%%%%%%%%%%%%%%%%%%%%%%%%%%%%%%%%%%%

%%%%%%%%%%%%%%%%%%%%%%%%%%%%%%%%%%%%%%%
% Related Work
%%%%%%%%%%%%%%%%%%%%%%%%%%%%%%%%%%%%%%%
\section{Related Work}

The bilateral filter (BF) was rediscovered several times by Aurich and Weule \cite{Aurich:1995}, Smith and Brady \cite{Smith:1997}, who introduced the SUSAN filter, and Tomasi and Manduchi \cite{Tomasi:1998} who gave BF its name. It was later popularized by Durand and Dorsey \cite{Tomasi:1998}. For a recent survey of BF see \cite{Paris:2009}. \cite{KiefelJG15} learns a high dimensional linear filter. Thus it generalizes the bilateral filter, which can be viewed as a Gaussian filter in high dimensions. 

The BF is just one of a large number of edge-preserving filters that include Anisotropic Diffusion \cite{Perona:1990}, guided image filter \cite{He:2010}, or the domain transform filter \cite{Gastal:2011} to name a few. These filters smooth images by averaging neighboring pixels. The weights are determined based on similarity in appearance and proximity in location. Correctly determining these weights determines what parts of the image should be smoothed and where smoothing should stop.

Joint/Cross BF \cite{Petschnigg:2004,Eisemann:2004} recovers weights on one image and applies it to another image. This concept was taken one step further with the guided image filter \cite{He:2010} where an image is assumed to be a locally linear model of the guidance image. 

The Rolling Guidance Filter \cite{Zhang:2014} uses the guidance image in a novel way leading to a scale-aware filter. That filter can be tuned to smooth out image structure at a particular scale by successively applying the BF with a properly selected guidance image. 

\cite{Li2016} solves a joint filtering problem. Instead of pre-designing the filter, it trains two CNNs that extract features from the filtered and the guidance images.

The WLS method \cite{FarbmanFLS08} treats edge preserving filtering as a weighted least square problem where the goal is to approximate the input image anywhere, except at sharp edges. The Euclidean distance in WLS can be replaced by the diffusion distance \cite{Farbman2010DME}. The diffusion distance between two points equals the difference between the probabilities of random walkers to start at both points and end up in the same point. To approximate this, \cite{Farbman2010DME} uses the dominant eigenvectors of the affinity matrix, dubbed diffusion maps. Diffusion maps can be efficiently calculated using the Nayst{\"o}m method.

Similarly to WLS, $L_0$ smoothing \cite{l0smoothing2011} approximates the input image with a piecewise constant image by controlling, through $L_0$ regularization, the number of edges allowed at the output image.
Xi {\em et al.} \cite{Xu2012} assumes that an image is composed of structure and texture. Their goal is to separate the two. To achieve that, they measure the relative total variation per patch and use it as a smoothness term for an optimization problem. 

In the field of edge detection there has been great progress in recent years. This progress can be quantitatively measured on the Berkeley Segmentation Data Set \cite{MartinFM04}. Some of the leading methods include Normalized Cuts and its derivative work \cite{ShiM00,ArbelaezMFM11} that treat the problem as a spectral clustering problem where affinities between pixels are trained offline. Structured Edge Detector \cite{DollarZ15} trains a structured random forest on a large training set and then applies it to detect true edges in the query image. 

Semantic filtering \cite{Yang_2016_CVPR}, uses the edges of \cite{DollarZ15} to modify the distances in the transformed domain of \cite{Paris:2009}. It does so by re-weighting the distance between neighboring pixels according to its confidence in the edge between them. We, in contrast, rely on pixel co-occurrences. This gives us the freedom to determine from where to learn co-occurrences.

Co-occurrences were recently used for boundary detection \cite{Isola:2014}. They collect co-occurrence statistics (termed Pointwise Mutual Information, or PMI, in their paper) to learn the probability of boundaries in an image and use that information to compute the affinities required by spectral clustering. The method performs very well on the Berkeley Segmentation Data Set \cite{MartinFM04}. 

Co-occurrence information was first introduced by Haralick {\em et al.} \cite{Haralick:1973}. They proposed 14 statistical measures that can be extracted from the co-occurrence matrix and be used to measure similarity between textures. Later, color correlograms, that also rely on co-occurrence data, were used by Huang {\em et al.} \cite{Huang:1997} as image descriptor within an image retrieval system. Finally, co-occurrence statistics was also used with graph cuts by Ladicky {\em et al.} \cite{Ladicky:2013} where the goal was to solve a label assignment problem such that the labels will satisfy some given co-occurrence matrix.

\section{Co-occurrence Filter}

Linear filters take the form:
\begin{equation}
J_p = \frac{\sum_{q \in N(p)}{w(p,q) \cdot I_q}}{\sum_{q \in N(p)}{w(p,q)}}
\label{eq:linear_filter}
\end{equation}
where $J_p$ and $I_q$ are output and input pixel values, $p$ and $q$ are pixel indices, and $w(p,q)$ is the weight of the contribution of pixel $q$ to the output of pixel $p$. We consider gray scale images for now. Color images will be discussed later.

In Gaussian filter, $w(p,q)$ takes the form of:
\begin{equation}
w(p,q) = exp(-\frac{d(p,q)^2}{2\cdot{\sigma_s}^2}) \triangleq G_{\sigma_s}(p,q)
\end{equation}
where $d(p,q)$ is the Euclidean distance, in the image plane, between pixels $p$ and $q$, and $\sigma_s$ is a user specified parameter. Since $w(p,q)$ does not depend on image content, the filter is shift invariant.

In the Bilateral filter, $w(p,q)$ takes the form of:
\begin{equation}
w(p,q) = G_{\sigma_s}(p,q) \cdot exp(-\frac{|I_{p}-I_{q}|^2}{2\cdot{\sigma_r}^2})
\end{equation}
where $\sigma_r$ is a user specified parameter and in this case $w(p,q)$ depends on image content and the filter is shift-variant.

\subsection{Definition}
We define the Co-occurrence filter to be: 
\begin{equation}
J_{p} = \frac{\sum_{q \in N(p)}{G_{\sigma_s}(p,q) \cdot M(I_{p},I_{q}) \cdot I_{q} } }{\sum_{q \in N(p)}{{G_{\sigma_s}(p,q) \cdot M(I_{p},I_{q})}}}
\label{eq:cof}
\end{equation}
Which means that $w(p,q)$ takes the form of:
\begin{equation}
w(p,q) = G_{\sigma_s}(p,q) \cdot M(I_{p},I_{q})
\label{eq:weight_cof}
\end{equation}
where $M$ is a $256 \times 256$ matrix (in the case of the usual gray scale images) that is given by:
% This definition can work with any any real valued matrix $M$, and we will get to this later. For now, we will focus on a specific way to compute $M$ from co-occurrence data. Formally:
\begin{equation}
M(a,b) = \frac{C(a,b)}{h(a)h(b)}.
\label{eq:M}
\end{equation}
In words, $M(a,b)$ is based on the co-occurrence matrix $C(a,b)$ that counts the co-occurrence of values $a$ and $b$ divided by their frequencies (i.e., the histogram of pixel values), $h(a)$ and $h(b)$, in the image. By construction, $M$ is symmetric. To prevent division by zero we add a small constant to the denominator. Formally:
\begin{equation}
C(a,b) = \sum_{p,q} exp(-\frac{d(p,q)^2}{2\cdot{\sigma}^2}) [I_{p}=a] [I_{q}=b]
\label{eq:Cab_hard}
\end{equation}
and
\begin{equation}
h(a) = \sum_{p} [I_{p}=a]
\label{eq:Ca}
\end{equation}
where $\sigma$ is a user specified parameter and $[\cdot]$ equals $1$ if the expression inside the brackets is true and $0$ otherwise.

The co-occurrence matrix integrates all co-occurrences across all distances, weighted by their distance, in the image plane. This weight captures our belief that co-occurrences that occur far away carry a lower weight. In theory, we should sample all pixel pairs in the image plane. In practice, we consider only pixel pairs within a window. This differs from the usual gray-level co-occurrence matrix (e.g., \cite{Haralick:1973}) that is defined for a particular distance between pairs of pixels. 

%If $C(a,b)$ and $h(a)$ are normalized to be probability distribution functions and we take the $log$ of $M(a,b)$, we get the Pointwise Mutual Information (PMI). In practice, we did not observe noticeable difference between using $M(a,b)$ as defined above compared to PMI, hence, for the rest of the paper we will stick to the normalized co-occurrence definition. 

\begin{figure}[t]
\begin{center}
\setlength{\tabcolsep}{1.5pt}
\renewcommand{\arraystretch}{1}
\begin{tabular}{c c c c}
\includegraphics[width= 0.29\linewidth]{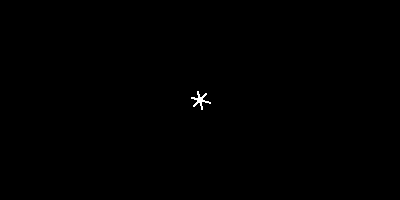}& 
\includegraphics[width= 0.29\linewidth]{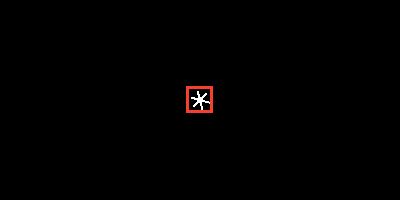} & 
\includegraphics[width= 0.29\linewidth]{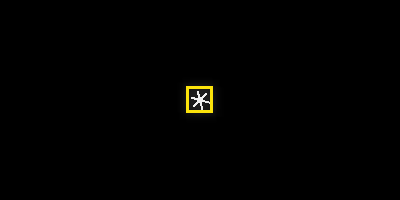} & 
\includegraphics[width= 0.073\linewidth]{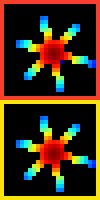} \\

\includegraphics[width= 0.29\linewidth]{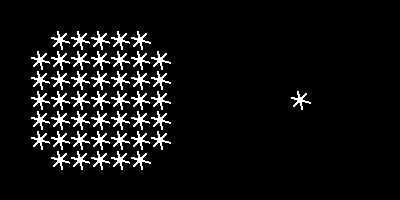}& 
\includegraphics[width= 0.29\linewidth]{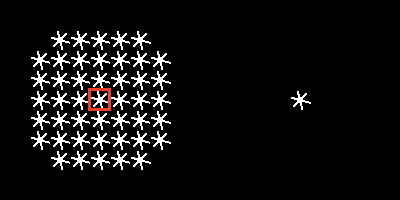} & 
\includegraphics[width= 0.29\linewidth]{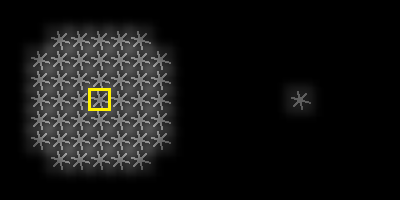} & 
\includegraphics[width= 0.073\linewidth]{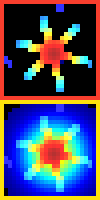} \\
\begin{comment}
\includegraphics[width= 0.29\linewidth]{multi_in.png}& 
\includegraphics[width= 0.29\linewidth]{multi_bf_weight.png} & 
\includegraphics[width= 0.29\linewidth]{multi_res.png} & 
\includegraphics[width= 0.073\linewidth]{multi_zoom.png} \\
\end{comment}

(a)& (b) & (c) & (d)  \\ 

\end{tabular}
\end{center}

\caption{ \textbf{ Role of context in CoF:} (a) input image, (b) BF, (c) CoF, (d) Zoom ins.
The BF filters the top and bottom images the same way. The CoF, on the other hand, filters them differently. The zoom ins shows the weight assigned to pixels when filtering the center pixel. Observe how the weights of CoF change depending on the content of the image.}
\label{fig.stars}
\end{figure}

\subsection{Properties}

Analyzing equation~\ref{eq:Cab_hard}, we observe that when $\sigma$ goes to $0$, $C(a,b)$ converges to a diagonal matrix. This is because the weight for every pair of pixels $p$ and $q$ goes to zero, except for the case $p=q$. Plugging this back into equation~\ref{eq:M} we have that $M$ is also a diagonal matrix, with elements on the diagonal taking the form:
\begin{equation}
M(a,a) = \frac{C(a,a)}{h(a)h(a)} = \frac{h(a)}{h(a)^2} =\frac{1}{h(a)}
\end{equation}
As a result, CoF becomes a delta function that does not change the input image at all. This is because each pixel is only affected by pixels with the {\em same} intensity value. 

At the other extreme, when $\sigma$ goes to $\infty$, then $C(a,b)=h(a)h(b)$. This is because the weight is equal for all pairs of pixels $p$ and $q$, and $C(a,b)$ is simply the product of the frequencies of values $a$ and $b$. Plugging this back into equation~\ref{eq:M} we have that:
\begin{equation}
M(a,b) = \frac{C(a,b)}{h(a)h(b)} = \frac{h(a)h(b)}{h(a)h(b)} = 1.
\end{equation}
That is, the matrix $M$ converges to the all one matrix, and the CoF becomes the Gaussian filter.
The bilateral filter can be constructed manually as a band-diagonal matrix $M$.

\begin{figure}[t]
\begin{center}
\setlength{\tabcolsep}{1.5pt}
\renewcommand{\arraystretch}{1}
\begin{tabular}{c c c c c}

\includegraphics[width= 0.22\linewidth]{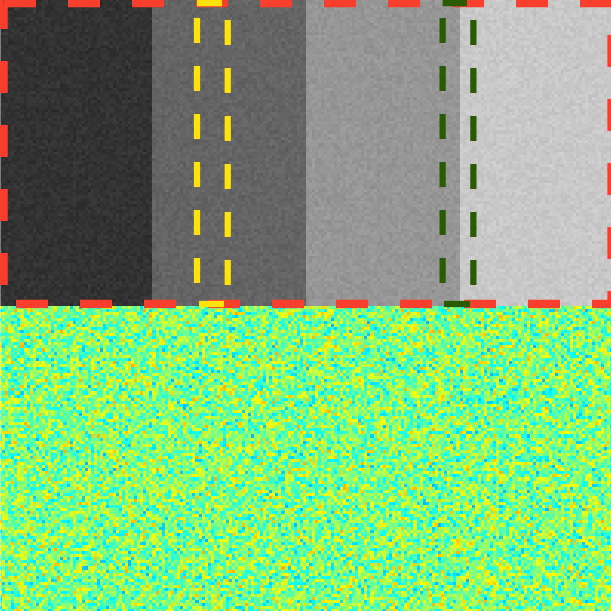}& 
\includegraphics[width= 0.22\linewidth]{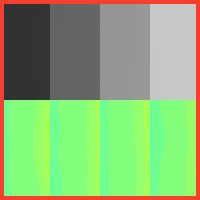} &
\includegraphics[width= 0.22\linewidth]{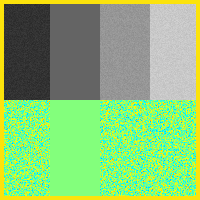}&
\includegraphics[width= 0.22\linewidth]{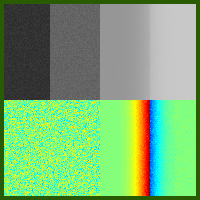} &
\includegraphics[width= 0.0733\linewidth]{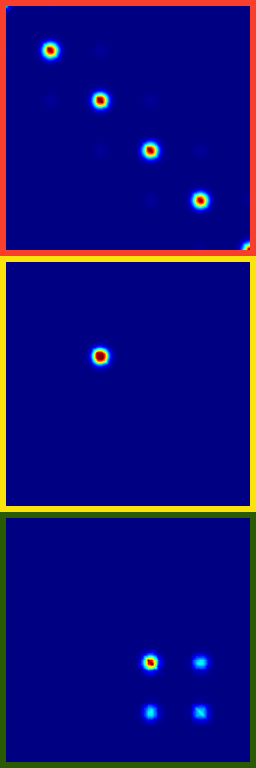}\\
(a) & (b) & (c) & (d) & (e)  \\

\end{tabular}
\end{center}
\caption{ \textbf{ Collecting co-occurrence statistics:} Co-occurrence statistics can be collected from different parts of the image. (a) input image contaminated with white Gaussian noise. Bottom part of the image shows difference between input image and clean image (not shown). (b) result of CoF when collecting statistics from all of the image (red dashed rectangle in (a)). (c) result of CoF when collecting statistics from one region (yellow dashed rectangle in (a)). That particular region is smoothed out, the rest of the image is not. (d) result of CoF when collecting statistics along the edge between two regions (green dashed rectangle in (a)). That particular edge between regions is smoothed out, as well as the two neighboring regions, the rest of the image is not. (e) The co-occurrence matrices corresponding to (b-d), from top to bottom, respectively.}
\label{fig.stripes}
\end{figure}

Figure~\ref{fig.stars} demonstrates the importance of context in CoF. The top row shows an image of a lone white star against a dark background. In this case, CoF and BF behave similarly. They preserve the sharp intensity difference between the white pixels of the star and the black pixels of the background. The bottom row show a galaxy of stars. The BF is completely agnostic to the presence of multiple stars in the image. CoF, on the other hand, behaves quite differently. Because there are multiple stars, the co-occurrence matrix picks up the co-occurrences of black and white pixels and the filtered image shows a milky result where black and white pixels are mixed. We emphasize that we did not change any of the parameters of CoF at all. Everything is dictated by the data.

Figure~\ref{fig.stripes} shows what happens when we pick different parts of the image from which to collect co-occurrence data. The input image consists of a sequence of step edges corrupted with some white Gaussian noise. Collecting co-occurrence data from all the image will lead CoF to filter out the noise with minimal smoothing of the step edges. Collecting co-occurrence statistics from part of one flat region will smooth all that region but will keep noise and sharp edges in other parts of the image intact. Finally, collecting co-occurrence statistics from the vicinity of a step edge will cause CoF to smooth out that particular step edge and the two neighboring regions.

\begin{figure}[t]
\begin{center}
\setlength{\tabcolsep}{1.5pt}
\renewcommand{\arraystretch}{1}
\begin{tabular}{c c c c}
\includegraphics[width= 0.22\linewidth]{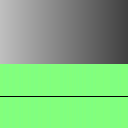} & 
\includegraphics[width= 0.22\linewidth]{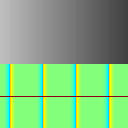} & 
\includegraphics[width= 0.22\linewidth]{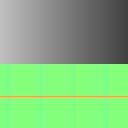} &
\includegraphics[width= 0.22\linewidth]{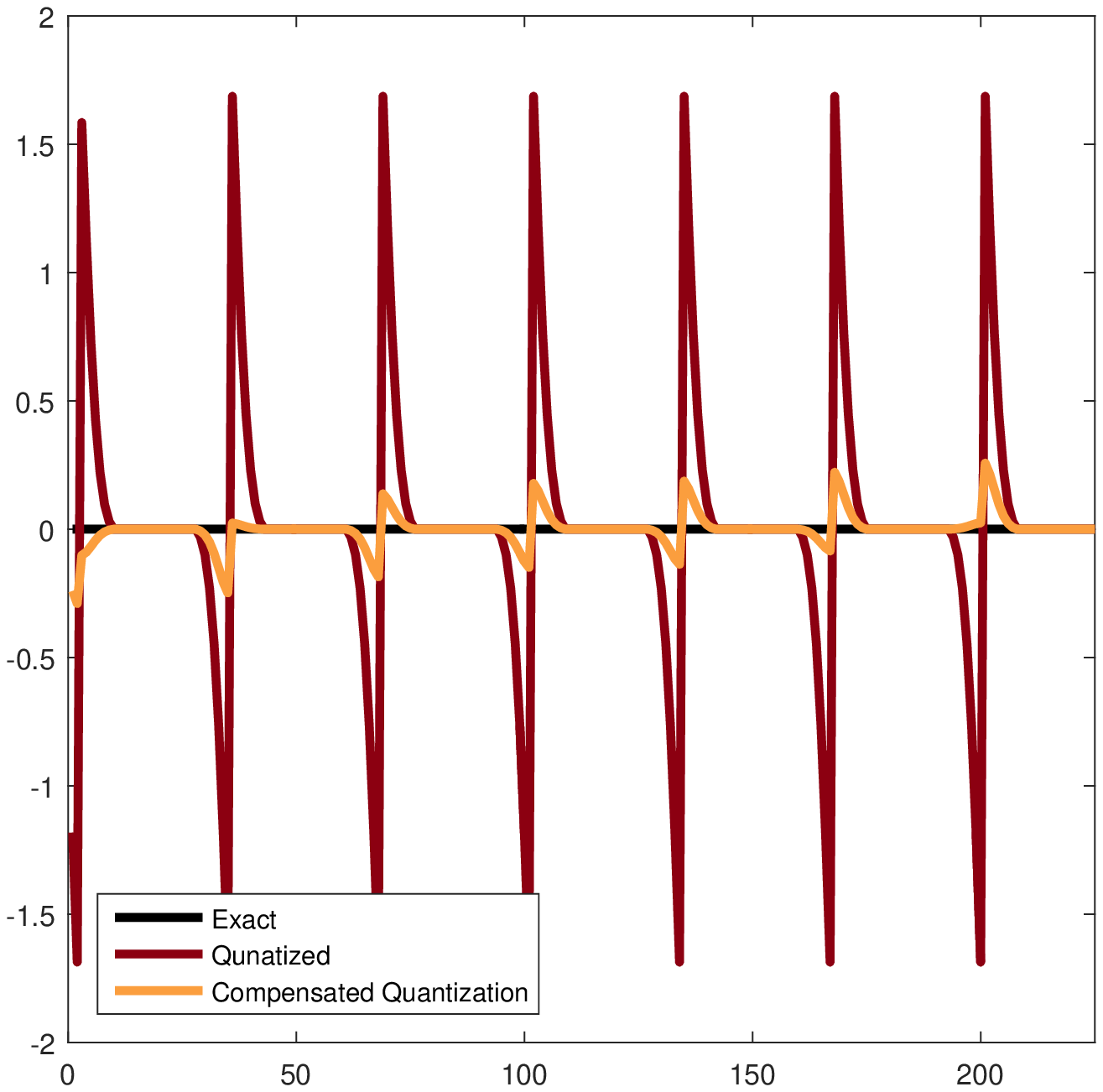} \\ 

(a) & (b) & (c) & (d) \\ 
\end{tabular}
\end{center}
\caption{\textbf{ Effects of quantization:} (a) CoF on image with $256$ gray values. (b) CoF on image with $32$ gray values with hard cluster assignment. (c) CoF on image with $32$ gray values with soft cluster assignment. (d) 1D profile of a particular row. The bottom half of each image shows the difference between the filtered image and the clean input image. The non-quantized result of (a) gives the exact solution. Soft quantization (c) gives better results than (b). See text for more details.}
\label{fig:quan}
\end{figure}
\subsection{Guided CoF}\label{seq:Guided_CoF}
So far we have assumed the input image is a gray scale image. We now extend CoF to work on color images. 

One can use equation (\ref{eq:Cab_hard}) to calculate co-occurrence in color space. This means constructing a $256^3 \times 256^3$ co-occurrence matrix. This matrix is too large for practical purposes. Moreover, the number of pixels in a typical image is too small to properly sample that space. We therefore quantize, using k-means, the pixel values of $I$ to produce a guidance image $T$. This solves both problems. The size of the co-occurrence matrix of $T$ is only $k \times k$, where $k$ is the number of quantized values, and the number of pixels in $T$ is enough to properly sample this space. 

Let $M_T$ denote the co-occurrence matrix of $T$. Then the guided CoF filter, $\mbox{CoF}(I,M_T)$, is given by:

\begin{equation}
J_{p} = \frac{\sum_{q \in N(p)}{G(p,q) M_{T}(T_{p},T_{q}) \cdot I_{p}}}{\sum_{q \in N(p)}{G(p,q) M_{T}(T_{p},T_{q})}}
\label{eq:quan_cof}
\end{equation}

The introduction of clustering changed equation~(\ref{eq:cof}) into equation~(\ref{eq:quan_cof}). We collect co-occurrence statistics from $T$ and use it to guide the filtering, hence we denote it the guidance image. This resembles the guided version of the bilateral filter, where the filtered image differs from the image that is used to compute color distances. The next subsection discusses how to collect $M_{T}$. 

\subsubsection{Quantized Co-occurrences}
Let $\{ \tau_l \}_{l=1}^{k}$ denote $k$ clusters after clustering pixel values. Then a straightforward way to extend equation~\ref{eq:Cab_hard} is to let:
\begin{equation}
%C_{hard}(\tau_{a},\tau_{b}) = \sum_{p,q} exp(-\frac{d(p,q)^2}{\sigma^2}) [p \in \tau_{a}] [q\in \tau_{b}]
C_{hard}(\tau_{a},\tau_{b}) = \sum_{p,q} exp(-\frac{d(p,q)^2}{2\cdot\sigma^2}) [T_{p} = a] [T_{q} = b]
\label{eq:Cab_hard_cluster}
\end{equation}
where $\tau_{a}$ and $\tau_{b}$ denote two clusters, and $T_{p} = a$ means that pixel $p$ belongs to cluster $\tau_{a}$. We term this approach hard clustering, because each pixel is assigned to its closest cluster center.

The time complexity of computing co-occurrences using hard clustering is $O(n \cdot r^{2})$, where $n$ is the number of pixels in the image and $r$ is the size of the window. This is because at each pixel location we must compute co-occurrence statistics for $r^2$ pixel pairs. In theory $r=n$, in practice we use a small window ($r=15\times15$).

However, such an approach introduces severe artifacts. These artifacts are created because pixel values that are nearby in the original space might be mapped to two different clusters in the quantization step.  

Figure \ref{fig:quan} illustrates the problem on a gray scale image. The left image shows a simple ramp image with gray scale values ranging from $0$ to $255$. Running CoF on it will leave the image unchanged because each intensity value co-occurs with the same number of intensity values above and below it. Collecting co-occurrence statistics using equation~\ref{eq:Cab_hard_cluster} introduces noticeable artifacts (see Figure~\ref{fig:quan}(b)).

To fix that, we relax the assignment of a pixel value to a single cluster. Instead, we use soft assignment. We assign a probability for the pixel value to belong to each of the clusters, using the following:

\begin{equation}
C_{soft}(\tau_{a},\tau_{b}) = \sum_{p,q} exp(-\frac{d(p,q)^2}{2\cdot\sigma^{2}}) Pr( p \in \tau_{a} ) Pr( q \in \tau_{b} )
\label{eq:Cab_soft_cluster}
\end{equation}

Unfortunately, moving from hard to soft assignment comes at a high computational cost. The cost of collecting co-occurrence statistics using soft assignment is $O(n \cdot r^{2} \cdot k^{2})$ operations, as opposed to the $O(n \cdot r^2)$ of hard assignment. 

To overcome this, let $Pr(p \in \tau) = K( I_{p}, \tau )$, where $K$ is a kernel function (i.e., a Gaussian):
\begin{equation}
K(a,b) = \frac{1}{Z}\exp(-\frac{||a - b||^{2}}{2\sigma_{r}^{2}})
\label{eq:kernel}
\end{equation}
for some user specified parameter $\sigma_{r}$ and normalization constant $Z$.

In words, $K$ measures the probability of assigning pixel $p$ to cluster $\tau$ based on the distance, in appearance space, between pixel value $I_p$ and cluster center $\tau$. We now make the approximation that $Pr(p \in \tau ) \approx K( \tau_{p}, \tau )$. That is, the distance between $I_p$ and $\tau$ is approximated by the distance between $\tau_{p}$ and $\tau$, where $\tau_{p}$ is the cluster center closest to $I_p$. In the supplemental, we derive the following relation:

\begin{equation}
C_{soft}(\tau_{a},\tau_{b}) \approx \sum_{k_{1},k_{2}} K(\tau_{a}, \tau_{k_{1}} ) \cdot K(\tau_{b}, \tau_{k_{2}} ) C_{hard}(\tau_{k_{1}}, \tau_{k_{2}} )
\label{eq:Cab_approx}
\end{equation}

The difference between Equation~\ref{eq:Cab_soft_cluster} and Equation~\ref{eq:Cab_approx} is that instead of working with all pixel values we only work with cluster centers. Using this approximation, the time complexity of collecting co-occurrence statistics using soft assignment drops from $O(n \cdot r^2 \cdot k^2)$ to $O(n \cdot r^{2} + k^{4})$. For a typical image of size $512 \times 512$ we have ($n=2^{18}$, $r=2^4 \times 2^4$, and $k=2^8$), which leads to a speed up of about $2^{10}=1024$, (i.e., three orders of magnitude). Figure~\ref{fig:quan}(c) shows the result of the soft assignment approach with our approximation. Observe how the staircase effect is greatly reduced. 

Algorithm~\ref{alg:cof} provides an outline of our method. In words, given a color image we first cluster its pixel values and use the quantized image, $T$, to calculate the hard quantization co-occurrence matrix, $C_{hard}$. This takes $O(n \cdot r^2)$. Once we have $C_{hard}$, we use cluster distances to approximate soft co-occurrence matrix, $C_{soft}$, that takes an additional $O(k^4)$. We divide $C_{soft}$ by the cluster probabilities and get the normalized co-occurrence matrix, $M_{T}$. Finally, we use this $M_{T}$ to filter the original image, $I$. 

\begin{algorithm}
\caption{Guided Co-occurrence Filtering}\label{alg:cof}
\renewcommand{\algorithmicrequire}{\textbf{Input image:}}
\renewcommand{\algorithmicensure}{\textbf{Filtered image:}}

\begin{algorithmic}[1]
\Require $I$
\Ensure $J$

\item {\footnotesize $[T, cc]$ $\leftarrow$ Quantize( $I$ )~~~~~~~~~~~~~~~~~~~~~~~~~~~~~~~\mbox{\% using k-means}}
\item {\footnotesize $C_{hard}$ $\leftarrow$ Compute Co-occurrence( $T$ )~~~~~\mbox{\% using equation}(\ref{eq:Cab_hard_cluster}) }
\item {\footnotesize $C_{soft}$ $\leftarrow$ Hard2Soft( $C_{hard}$, $cc$ )~~~~~~~~~~~~~~~\mbox{\% using equation}(\ref{eq:Cab_approx}) }
\item {\footnotesize $M_{T}$ $\leftarrow$ Cooc2PMI($C_{soft}$, $cc$)~~~~~~~~~~~~~~~~~~~~~\mbox{\% using equation}(\ref{eq:M}) }
\item {\footnotesize $J$ $\leftarrow$ CoF( $I, M_{T}$ ) ~~~~~~~~~~~~~~~~~~~~~~~~~~~~~~~~~~~~~~\mbox{\% using equation}   (\ref{eq:quan_cof}) }

\end{algorithmic} 
\end{algorithm}

\section{Results}

In this section we discuss some implementation details and demonstrate the performance of CoF. We conclude with two applications: background bluring and image recoloring.

Throughout this section we have used the guided version of the CoF filter, as described in section~\ref{seq:Guided_CoF}. For quantization, we use K-means, over lab colors, with $k=32$. To speed up the clustering we sample the image on a regular grid with a spacing of $10$ pixels in both rows and columns. For all the images, we collected co-occurrence over a window of $15 \times15$, with $\sigma_{s}^{2} = 2\cdot\sqrt(15)+1$. Unless explicitly mentioned, we used the same kernel for smoothing. Collecting co-occurrences takes about $2$ seconds for $1$ MP image. Filtering the image takes about $1.2$ seconds \footnote{Code will be released upon publication.}. All timing is for CPU implementation.

\begin{figure}
\begin{center}
\begin{tabular}{@{\hskip 0in}c@{\hskip 0.1in}c@{\hskip 0.1in}c@{\hskip 0.1in}c@{\hskip 0in}}
\includegraphics[width = 0.22\linewidth]{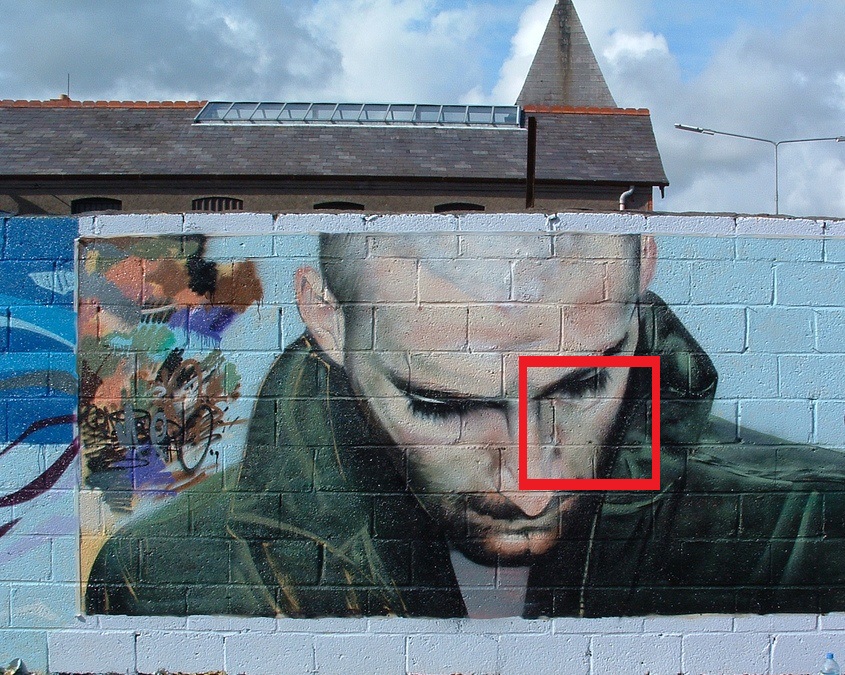} & 
\includegraphics[width = 0.22\linewidth]{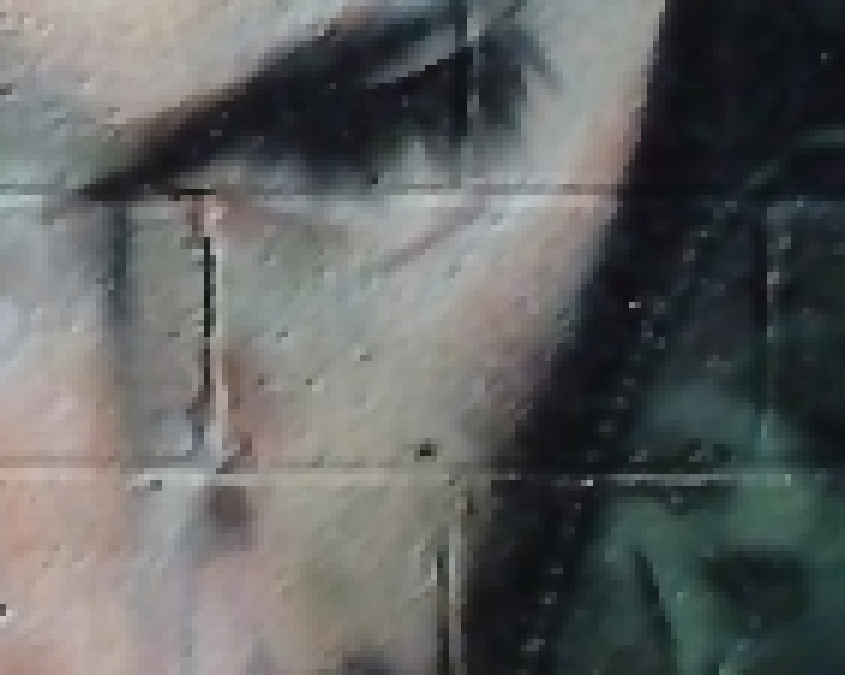} & 
\includegraphics[width = 0.22\linewidth]{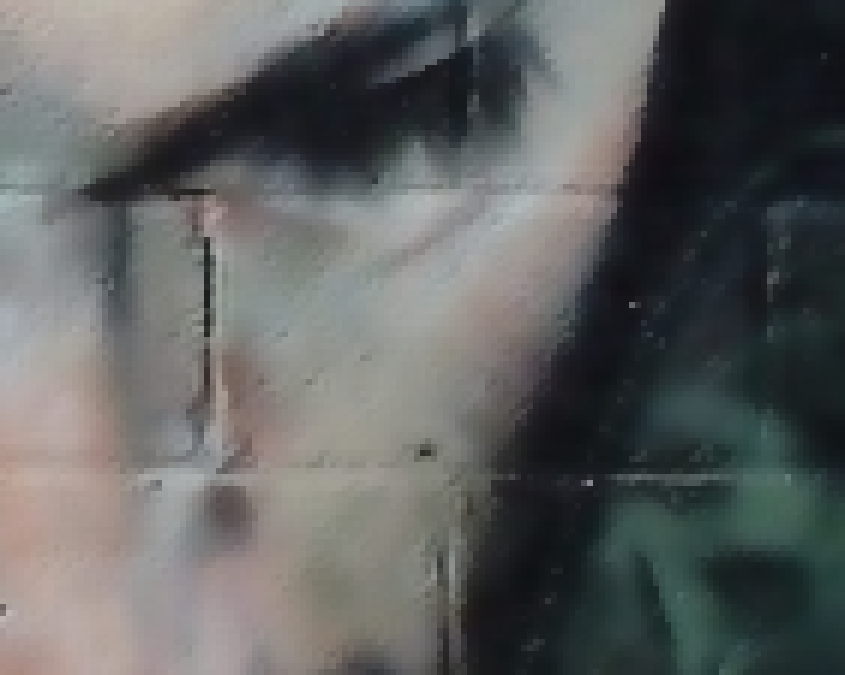} & 
\includegraphics[width = 0.22\linewidth]{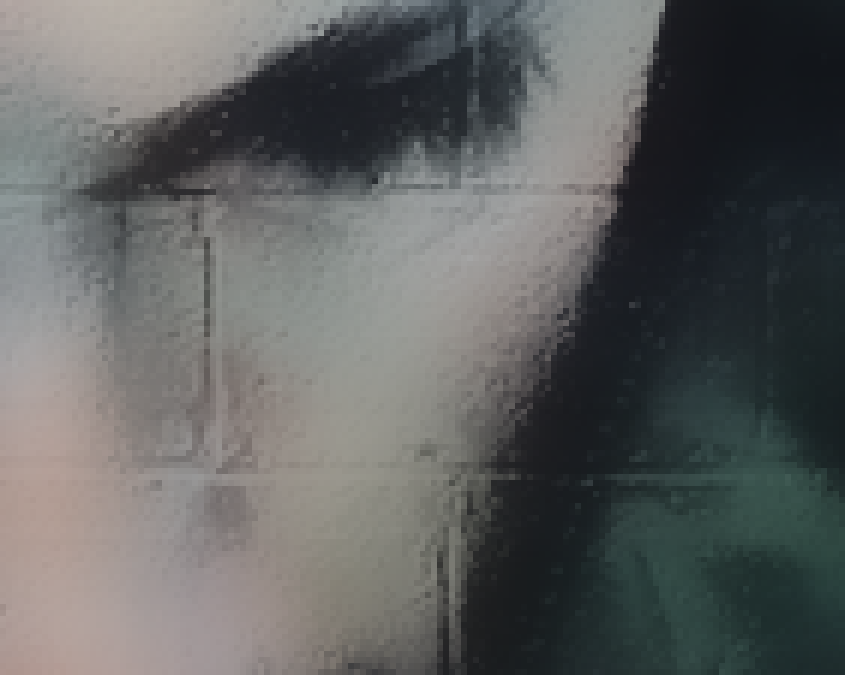} \\
input & $ws=3$ & $ws=5$ & $ws=15$\\
\end{tabular}
\end{center}

\caption{\textbf{ The effect of window size:} We show the effect of window size ($ws$), used when collecting co-occurrence statistics, on the behavior of CoF. The larger the window, the stronger the smoothing. The remaining edges stay sharp.}
\label{fig.graffiti}
\end{figure}
The first example, shown in Figure~\ref{fig.graffiti} shows the effects of the window size on the filter. As expected, the larger the window the larger are the objects that are smoothed by the filter. In all cases, though, the boundaries between textures remain sharp.

\begin{figure}
\begin{center}
\begin{tabular}{c@{\hskip 0.1in}c@{\hskip 0.1in}c@{\hskip 0.1in}c}
\includegraphics[width= 0.26\linewidth, frame]{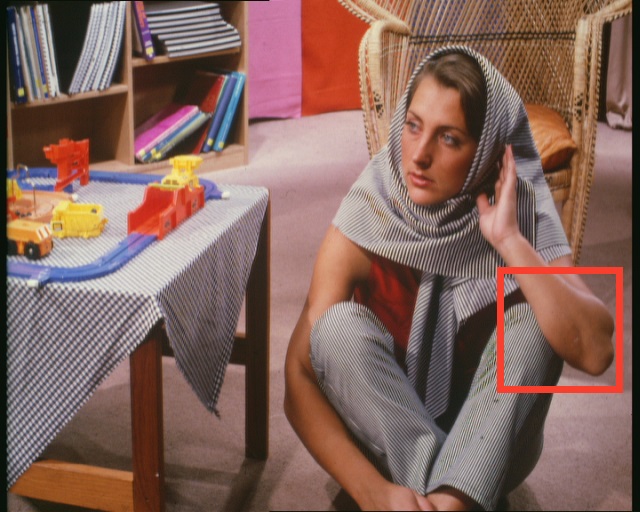} & 
\includegraphics[width= 0.21\linewidth]{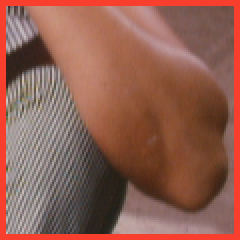} & 
\includegraphics[width= 0.21\linewidth]{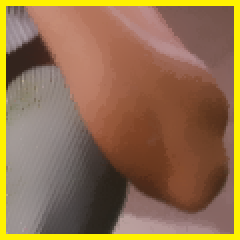} & 
\includegraphics[width= 0.21\linewidth]{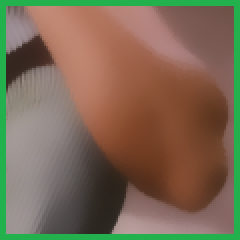} \\
Input &
Zoom in & 
Hard CoF & 
Soft CoF\\
\end{tabular}
\end{center}
\caption{{\bf The effect of soft quantization:} The red rectangle zooms in on Barbara's hand. The yellow rectangle zooms in on an image filtered with the hard clustering variant of the Co-occurrence filtered. The green rectangle zooms in on the soft clustering variant. Results are shown after 5 iterations of CoF. Notice how the quantization artifact on the hand disappear once we move to the soft version.}
\label{fig.Barbara}
\end{figure}
Figure~\ref{fig.Barbara} shows the importance of proper quantization. Working with hard clustering introduces strong quantization artifacts. Working with soft cluster assignment leads to a much smoother result. 

In Figure~\ref{fig.Barbara} we applied CoF multiple times. This raises the question: how to apply CoF iteratively? There are two ways to do that. Either by learning the co-occurrence statistics once, at the beginning, which we term Iterative CoF (I-CoF), or by learning the co-occurrence statistics after each round, which we term Rolling CoF (R-CoF). In Figure~\ref{fig.Barbara} we have used I-CoF. 

\begin{figure}
\begin{centering}
\begin{tabular}{@{\hskip 0in}c@{\hskip 0.05in}c@{\hskip 0.05in}c@{\hskip 0in}}
\includegraphics[width=0.566\linewidth]{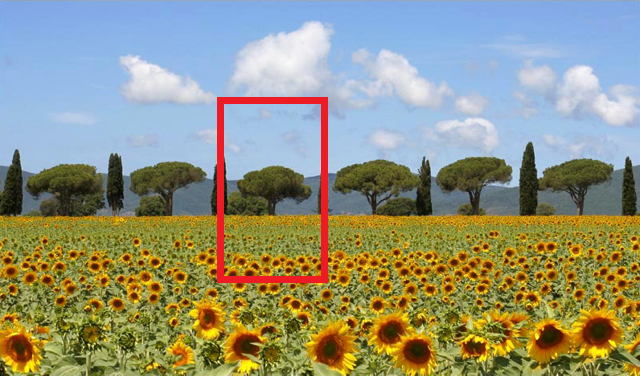} &
\includegraphics[width=0.19\linewidth]{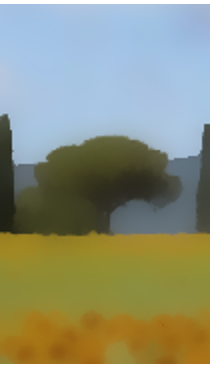} &
\includegraphics[width=0.19\linewidth]{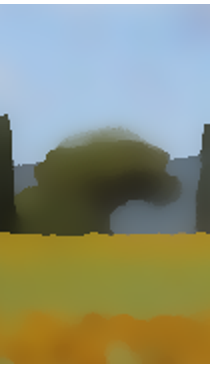} \\
{\small Input} & {\small Iterative} & {\small Rolling}\\
\end{tabular}
\end{centering}
\caption{\textbf{ Comparison of iterative vs. rolling CoF:} iterative (using the same $M$), rolling (updating $M$ after each iteration). }
\label{fig.i_vs_r}
\end{figure}
Figure~\ref{fig.i_vs_r} shows the difference between I-CoF and R-CoF after 10 iterations. As can be seen, I-CoF does a better job of smoothing texture while preserving sharp boundaries between textures. For the rest of this paper we use I-CoF when running CoF multiple times.

\begin{figure}
\begin{center}

\begin{tabular}{@{\hskip 0in}c@{\hskip 0.05in}c@{\hskip 0in}}
\includegraphics[width=0.48\linewidth]{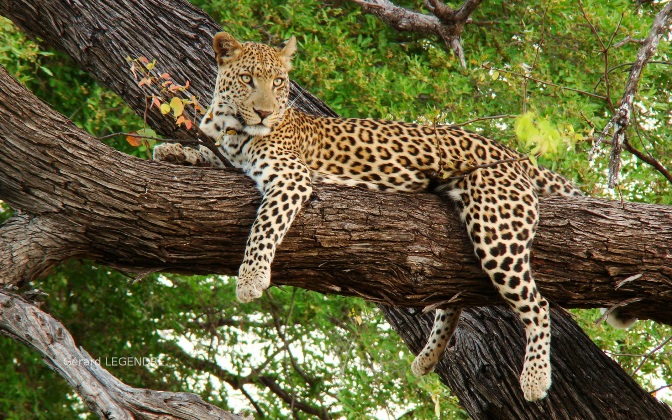} & 
\includegraphics[width=0.48\linewidth]{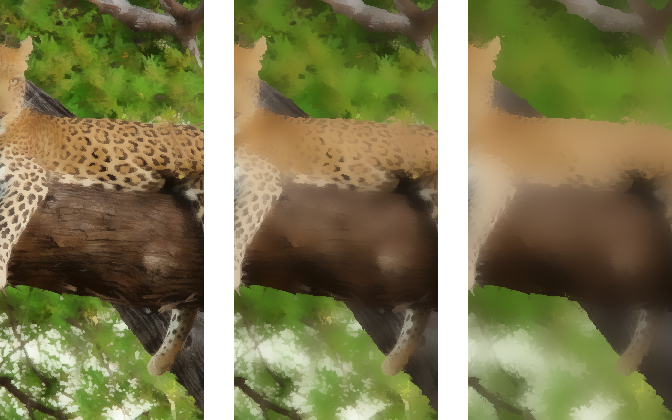} \\
\small{input} & \small{1, 3, and 10 iterations} \\
\multicolumn{2}{c}{ \includegraphics[width= 0.965\linewidth]{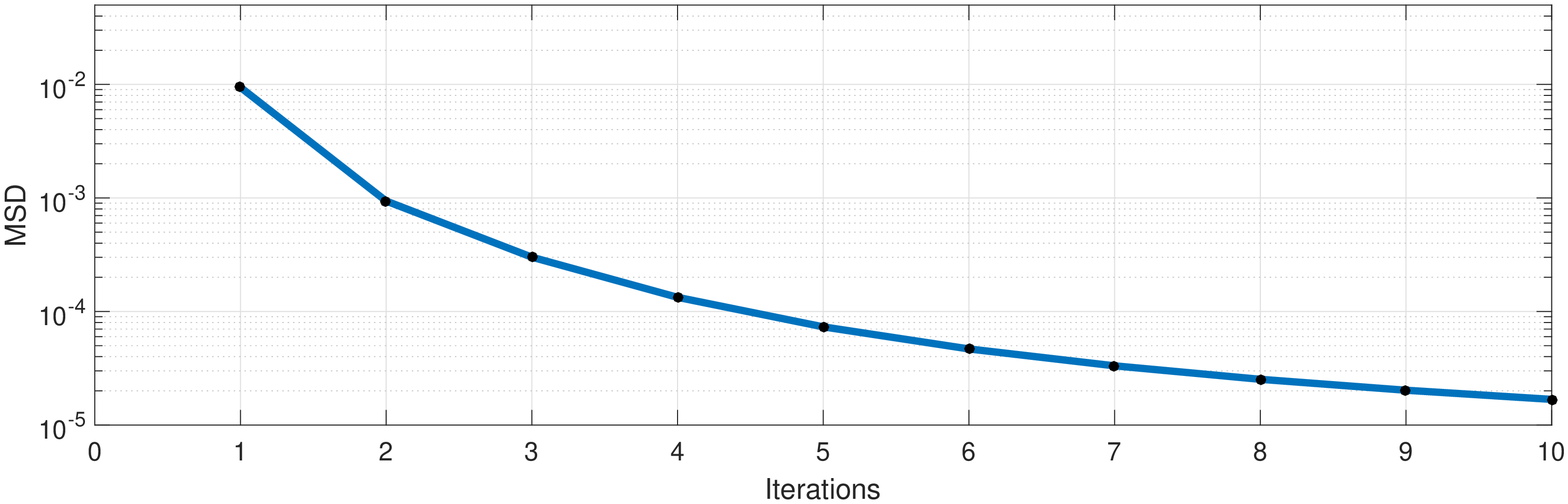} }

\end{tabular}
\end{center}
\caption{\textbf{Applying CoF Iteratively}: First row: shows the results of applying CoF for 1, 3 and 10 itterations. Bottom figure shows convergence rate (i.e., Mean-Squared-Difference, in intensity values, between successive iterations of the algorithm) on a semi-logarithmic scale.}
\label{fig.zebra}
\end{figure}

Figure~\ref{fig.zebra} shows the result of the algorithm after 1, 3 and 10 iterations. It also shows the mean of per-pixel squared differences between two iterations of the algorithm. As can be seen, the algorithm quickly converges.

\begin{figure*}%
\begin{center}
\begin{tabular}{@{\hskip 0in}c@{\hskip 0.05in}c@{\hskip 0.05in}c@{\hskip 0.05in}c@{\hskip 0in} }

\includegraphics[width=0.24\linewidth]{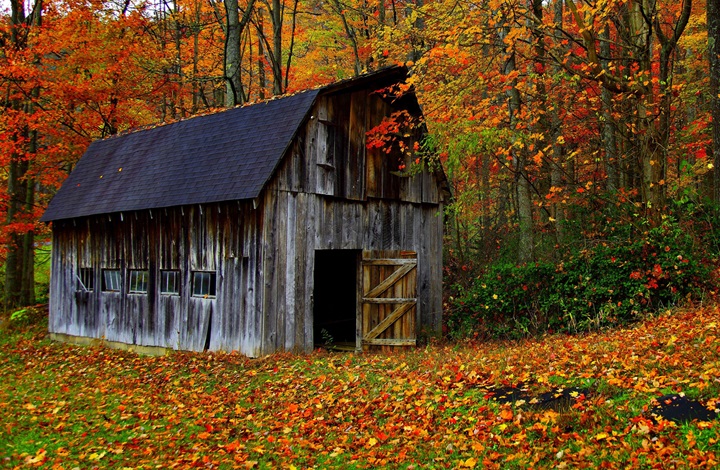}&
\includegraphics[width=0.24\linewidth]{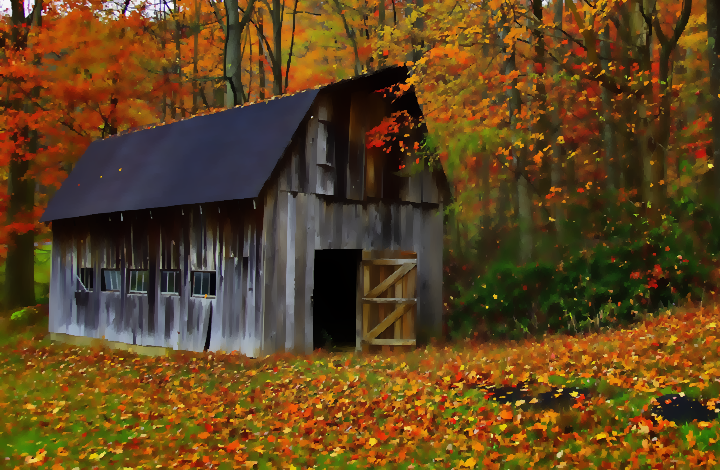}&
\includegraphics[width=0.24\linewidth]{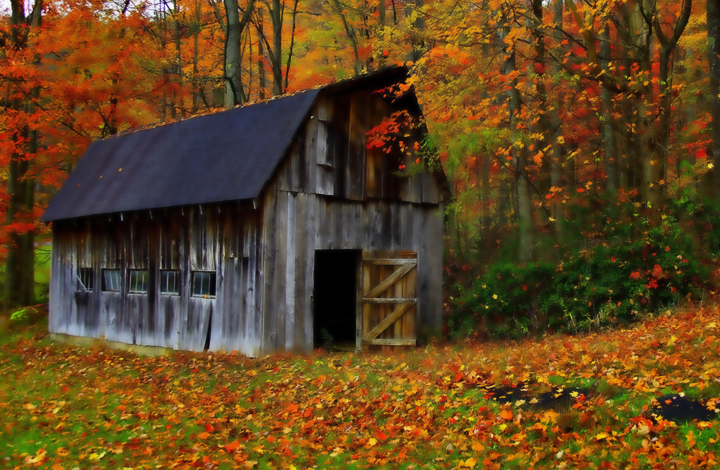}&
\includegraphics[width=0.24\linewidth]{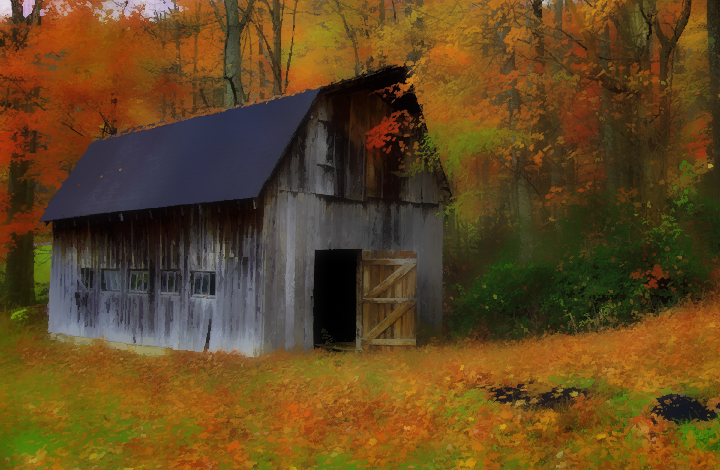} \\ 
{\small Input} & {\small Domain Transform~\cite{Gastal:2011}} & {\small Guided Image Filter~\cite{He:2010}} & \small{CoF}
\vspace{0.05in}\\ 

\includegraphics[width=0.24\linewidth]{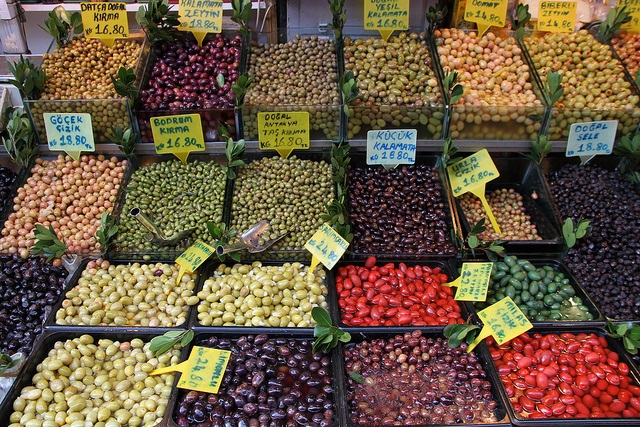}&
\includegraphics[width=0.24\linewidth]{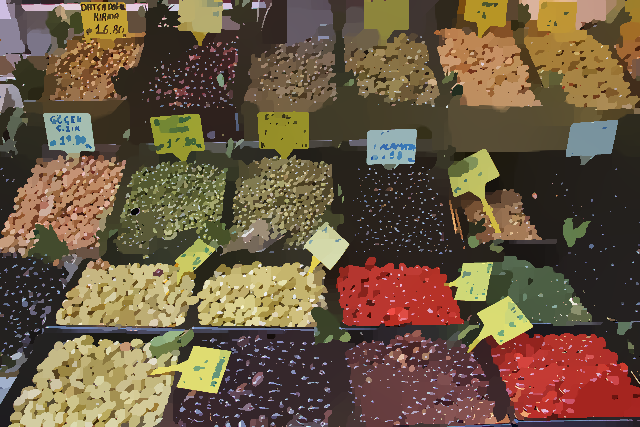}&
\includegraphics[width=0.24\linewidth]{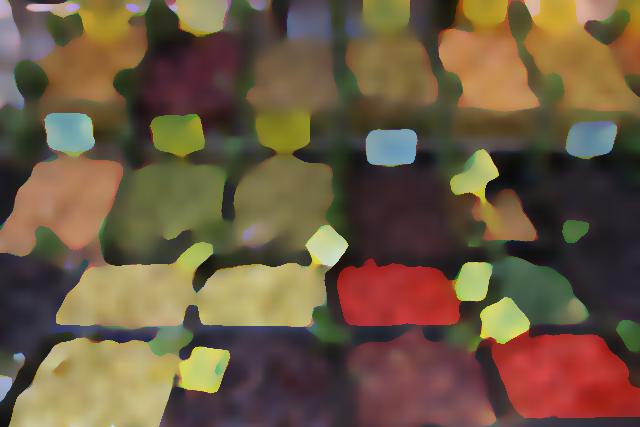}&
\includegraphics[width=0.24\linewidth]{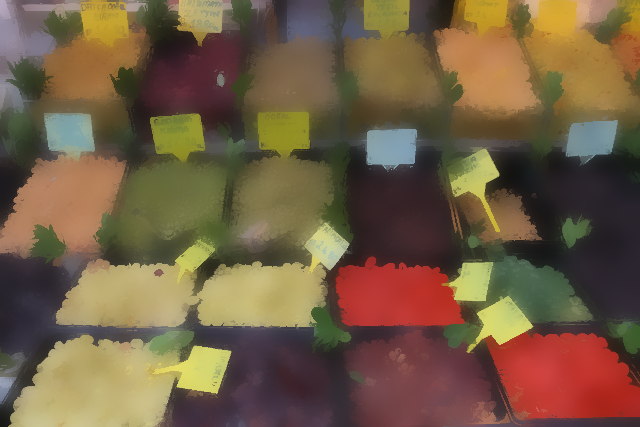}\\
{\small Input} & {\small $L_0$ Smoothing~\cite{l0smoothing2011}} & {\small Rolling Guided Filter~\cite{Zhang:2014}} & {\small CoF} 
\vspace{0.05in} \\

\includegraphics[width=0.24\linewidth]{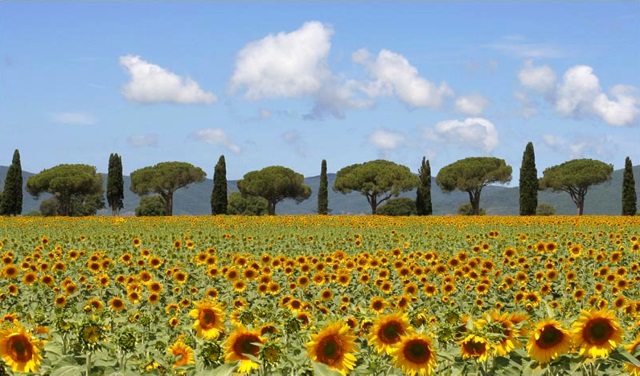}&
\includegraphics[width=0.24\linewidth]{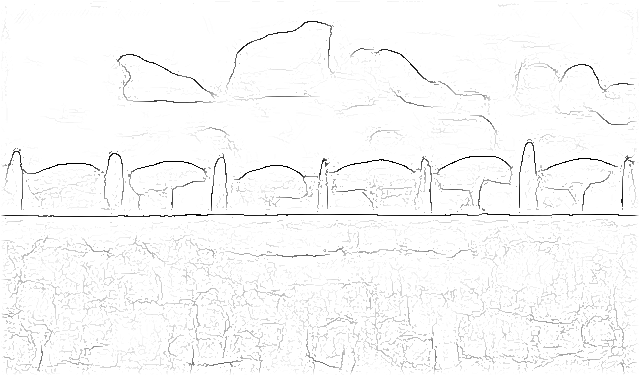}&
\includegraphics[width=0.24\linewidth]{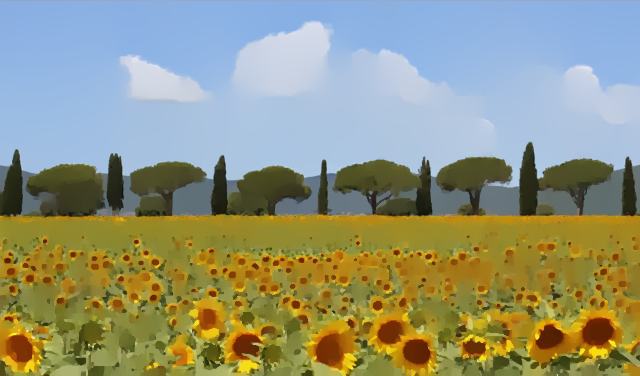}&
\includegraphics[width=0.24\linewidth]{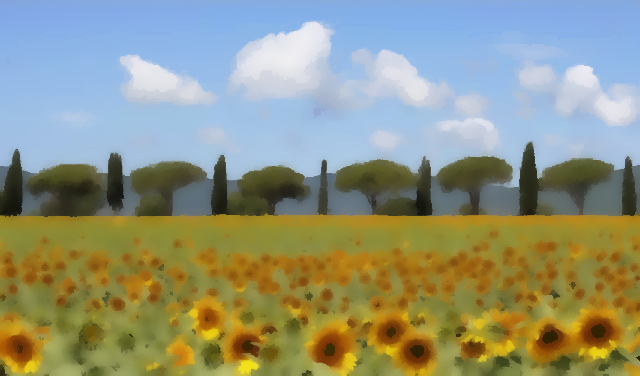} \\
{\small Input} & {\small Edge Map~\cite{DollarZ15}} & {\small Semantic Filter~\cite{Yang_2016_CVPR} } & {\small CoF} 
\vspace{0.05in}\\ 

\includegraphics[width=0.24\linewidth]{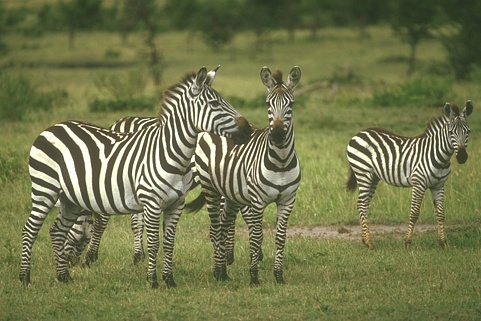}&
\includegraphics[width=0.24\linewidth]{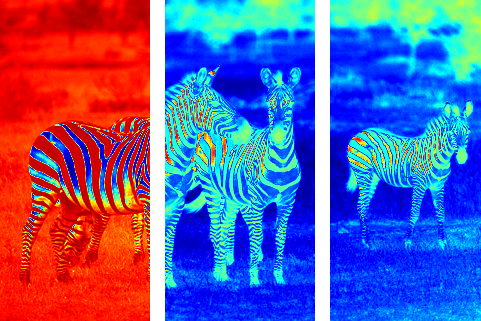}&
\includegraphics[width=0.24\linewidth]{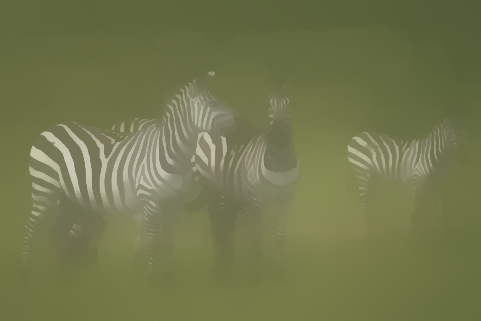} &
\includegraphics[width=0.24\linewidth]{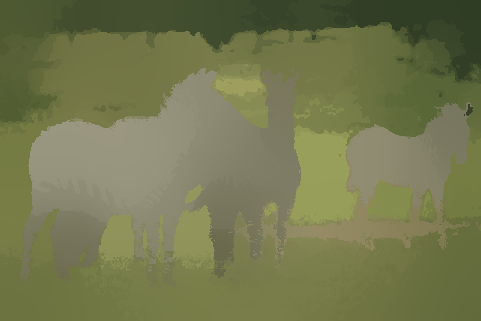} \\
{\small Input} & {\small $EV_{1}, EV_{0.65} \& EV_{0.18}$} & {\small WLS+DM~\cite{Farbman2010DME} } & {\small CoF}
\vspace{0.01in} \\

\end{tabular}
\end{center}
\caption{\textbf{Comparison to other Methods:} The first row compares CoF against Domain Transform~\cite{Gastal:2011} and Guided Image Filter~\cite{He:2010}. The second row against $L0$ smoothing~\cite{l0smoothing2011} and Rolling Guided Filter~\cite{Zhang:2014}. The third row against the Semantic Filter~\cite{Yang_2016_CVPR}. The last row against WLS with diffusion distances~\cite{Farbman2010DME}. }
\label{fig.rgf}
\end{figure*}

\begin{figure*}
\begin{center}
\vspace{0.15in}
\begin{tabular}{@{\hskip 0in}c@{\hskip 0.05in}c@{\hskip 0.05in}c@{\hskip 0.05in}c@{\hskip 0in}}
\includegraphics[width= 0.24\linewidth]{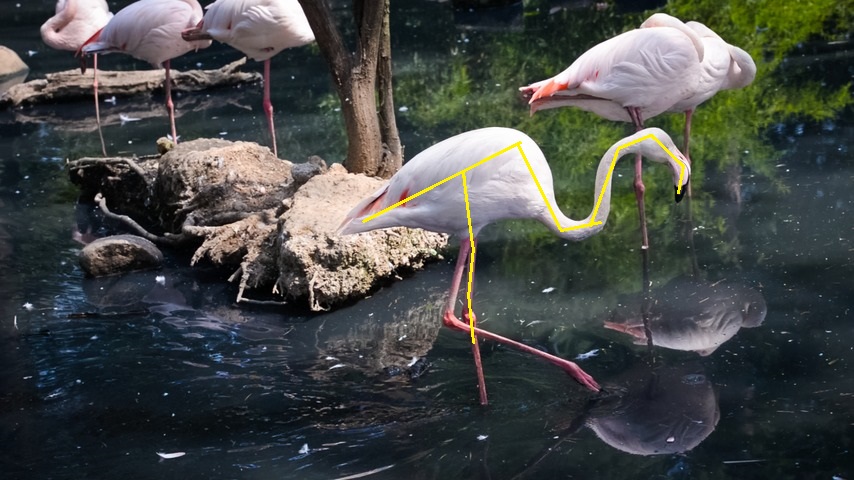}& 
\includegraphics[width= 0.24\linewidth]{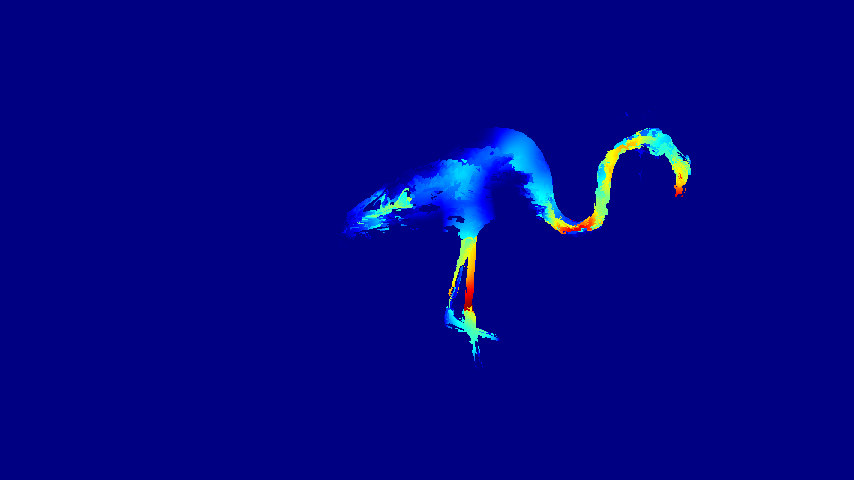} & 
\includegraphics[width= 0.24\linewidth]{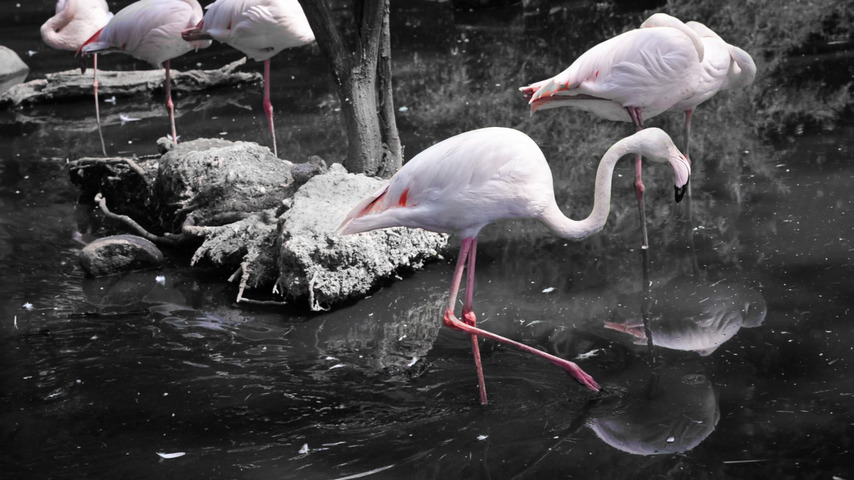} &
\includegraphics[width= 0.24\linewidth]{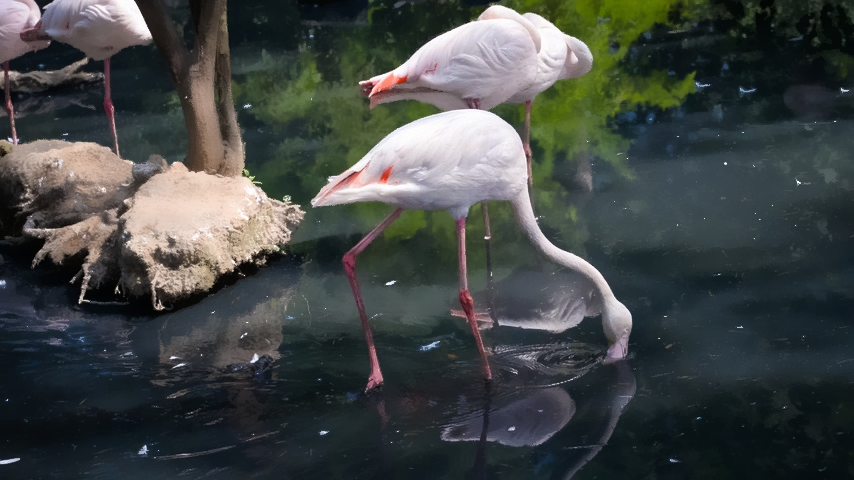}\\
{\bf (a) Input + Scribble} & {\bf (b) Mask} & {\bf (c) B\&W CoF} & {\bf (d) FB CoF 20 frames}\\

\includegraphics[width= 0.24\linewidth]{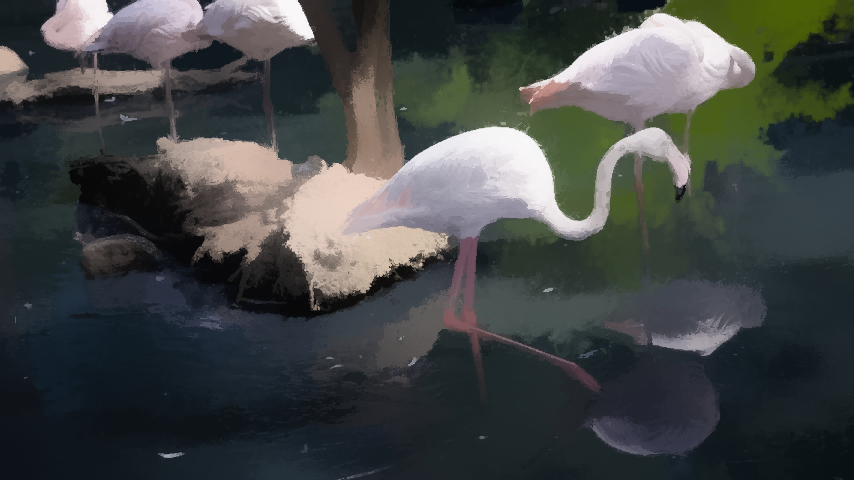}&  
\includegraphics[width= 0.24\linewidth]{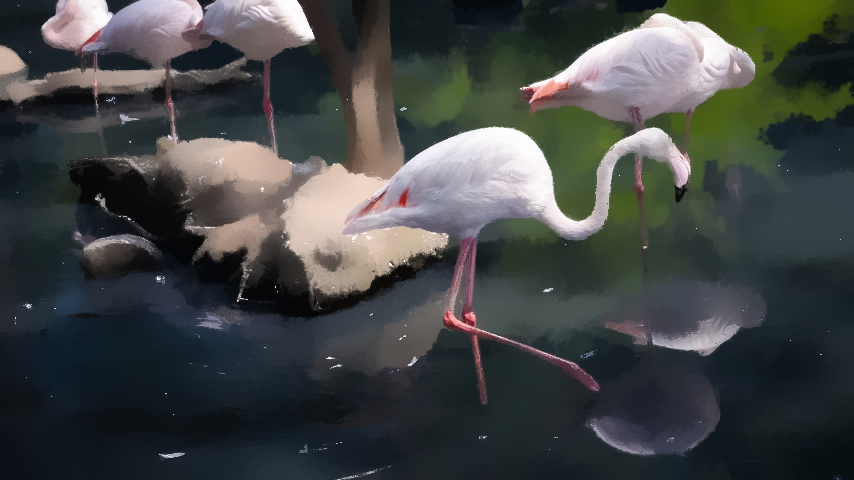}&
\includegraphics[width= 0.24\linewidth]{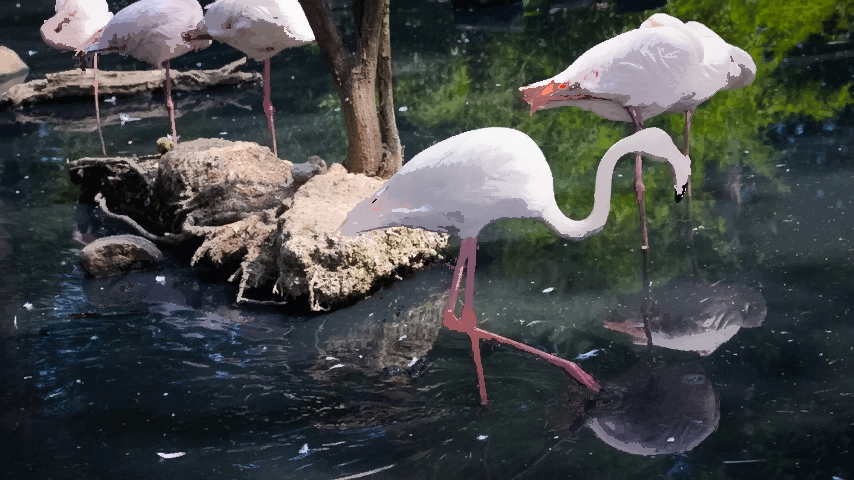}&
\includegraphics[width= 0.24\linewidth]{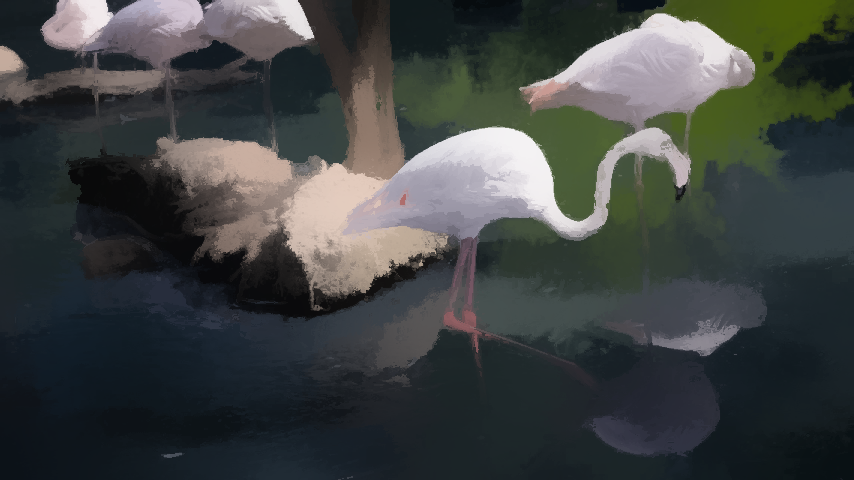}\\
{\bf (e) CoF} & {\bf (f) FB CoF} & {\bf (g) Foreground} & {\bf (h) Background}\\

\end{tabular}
\end{center}
\caption{ \textbf{Applications: A list of potential use cases for CoF. See text for details.} } 
\label{fig.example_flamingo}
\end{figure*}

Figure~\ref{fig.rgf} compares CoF with a number of edge preserving filters. Evaluating different filters is challenging because there is no agreed upon, objective error measure to optimize for. We therefore, resort to subjective evaluation, to illustrate the differences between CoF and each of them. On the first row we compare against Domain Transform \cite{Gastal:2011} and Guided Image Filter \cite{He:2010}. Both methods provide plausible results on the hut's roof. In addition, they enhance the leaves' colors, meaning red /green and yellow leaves will become a smoother red, green or yellow. This is due to the fact that both methods are edge preserving. CoF, on the other hand, learns that red, yellow and green are part of a texture and smooths them together. This doesn't come at the cost of smoothing across boundaries, see for example how nicely it preserves the sharp boundary between the hut and the leaves.

The second row compares CoF to $L_{0}$ smoothing \cite{l0smoothing2011} and the rolling guidance filter \cite{Zhang:2014}. $L_{0}$ performs a global smoothing operation that respects the strongest edges. One of its greatest applications is in image simplification. It works best for images with textures of modest gradients. If the image includes a texture with abrupt changes, take for example the pile of black olives in the center of the image, $L_{0}$ might wrongly respect some of the edges , and the texture will not be smoothed.  Rolling Guidance Filter (RGF) smooths texture up to a particular size. In this example we choose the window size to match the largest olive. Indeed it smoothed nicely all of the olives. However, that came at the price of rounding the price signs. In  addition, as the edges of the leaves that are located between the piles are smaller than the largest olive, RGF smoothed them out and damaged their structure. 

The third row compares CoF with Semantic filter \cite{Yang_2016_CVPR}. Semantic filtering uses the edge map produced by SED \cite{DollarZ15} to down-weight pixels that are not detected as edges. This produces great results inside textures with small gradients, see for example how nicely the trees are smoothed. However, in cases where the edge detector provide false edges, the semantic filter fails, see for example the artifacts in the middle of the sunflower field.

The forth row compares CoF to WLS enhanced with Diffusion distance \cite{Farbman2010DME}. We present the first three eignvectors. As can be seen, none of them cluster the black and white strips of the Zebra together. CoF, on the other hand, manages to fade the zebra into gray.

Figure~\ref{fig.example_flamingo} shows a number of potential applications of CoF. See supplemental for more examples, details and comparisons.

Figure~\ref{fig.example_flamingo}(a) shows the input image taken from a short video clip \cite{eth_biwi_01272}. On top of it, we show user supplied scribbles. Figure~\ref{fig.example_flamingo}(e) shows the result of running CoF on the image where the co-occurrence matrix was collected over the entire image, without using the scribbles. 

We next show how to use CoF for selective smoothing. To do that, we first need to convert the scribbles into a mask, so that we can collect sufficient statistics for the foreground co-occurrence matrix $M_F$ and the background co-occurrence matrix $M_B$. 

One way to do that is to use interactive segmentation. Instead, we show a different approach that is based solely on CoF. Let $S$ denote the sparse scribble image and compute the mask $L=\mbox{CoF}(S, M_T)$. Figure \ref{fig.example_flamingo}(b) shows the filtered result, which we threshold to get the foreground mask. This works because the scribble pixels belong to the foreground and the co-occurrence matrix, computed from $T$, makes CoF mix them. Note that the mask $L$ is not a perfect segmentation of the image. It might have mislabeled pixels, but the goal of this step is simply to extend the support of the scribbles. We found that it is better to miss a few foreground pixels than include background pixels that will distort the co-occurrence statistics. Once we have the mask $L$ we collect $M_F$ and $M_B$. Figures~\ref{fig.example_flamingo}(g) and ~\ref{fig.example_flamingo}(h) show the result of running CoF on $I$ with either $M_F$ or $M_B$.

A better control of the result can be achieved by properly combining $M_F$ and $M_B$. This is shown in Figure~\ref{fig.example_flamingo}(f). It was generated using the following filter:
\begin{equation}
J_p = \frac{\sum_{q \in N(p)} ( M_{F}(I_{p},I_{q}) \cdot I_p + M_{B}(I_{p},I_{q}) \cdot I_q )}{\sum_{q \in N(p)} ( M_{F}(I_{p},I_{q}) + M_{B}(I_{p},I_{q}) )} 
\label{eq:fb_cof}
\end{equation}

In words, if $p$ is a foreground pixel, then $M_{F}(p,q) >> M_{B}(p,q)$, for most of its neighbors, $q$. In this case, most neighbors contribute $I_p$, hence the resulting value, $J_{p}$, will remain close to $I_{p}$. This will keep the image sharp at foreground pixels. On the other hand, if $p$ is a background pixel, then $M_{F}(p,q) << M_{B}(p,q)$ for most of neighbors. This time, each neighbor contributes $I_{q}$, and the resulting value would be a weighted average of these values (i.e., smoothing). It is important to emphasize that the proposed algorithm might smooth foreground pixels slightly but background pixels will be smoothed much more.

Figure~\ref{fig.example_flamingo}(c) shows how to turn the background into grayscale, while keeping the object in full color. To do that, we use the following filter:

\begin{equation}
J_p = \frac{\alpha I_p^{color} + \beta I_p^{gray} }{\alpha + \beta} 
\label{eq:fb_gray}
\end{equation}

where 
\begin{equation}
\alpha = \sum_{q \in N(p)} M_{F}(I_{p},I_{q}),~~~
\beta = \sum_{q \in N(p)} M_{B}(I_{p},I_{q}).
\end{equation}

and $I^{gray}$ is a grayscale version of the image. Intuitively, $\alpha$ measures how well the neighboring pixels of pixel $p$ co-occur with it, under the foreground co-occurrence matrix $M_F$. Similarly, $\beta$ measures how well the neighboring pixels of $p$ co-occur with it, under the background co-occurrence matrix $M_B$. As a result, foreground pixels will prefer the $I^{color}$ while background pixels will prefer $I^{gray}$. 

The last example, shown in Figure~\ref{fig.example_flamingo}(d), shows how to use CoF in video. In this case the co-occurrence matrices collected on image~\ref{fig.example_flamingo}(a) can be applied to an image that is 20 frames apart in the video. Evidently, the learned co-occurrence matrices produce reasonable results.  

Taken together, Figure~\ref{fig.example_flamingo} shows the many ways CoF can be used to achieve various artistic results. %Here we focus on the use of CoF on its own but in the future we plan to investigate ways to combine CoF with other tools reported in the literature.

\section{Conclusions}
We proposed Co-occurrence Filter (CoF), a boundary preserving filter. CoF collects co-occurrence statistics from the image before applying the filter. A high co-occurrence weight causes pixel values to mix, leading to smoothing within textured region. On the other hand, low co-occurrence weight prevents pixels from mixing, leading to sharp boundaries between textured regions. We defined the filter, demonstrated its features and showed how it should be applied to color images.
We show results on various images and suggested several use cases that include learning co-occurrence statistics on parts of the image, or learning them on one image and applying it to another. Finally, we presented several use cases to demonstrate its power and potential.

\section*{Acknowledgement}
Part of this research was supported by ISF grant 1917/2015

\section*{Appendix}
We derive the connection between the co-occurrence matrix using hard and soft clustering. The former is faster to compute, but the latter is more accurate. We suggest an approximation that maintains the speed of the hard clustering approach with the visual quality of soft clustering.
Recall that calculating co-occurrence matrix using hard and soft assignments is given by: 

\begin{equation} 
C_{hard}( \tau_{a}, \tau_{b} ) = \sum_{i,j} exp(-\frac{d_{ij}^2}{\sigma^2}) [i \in \tau_{a}] [j\in \tau_{b}]
\label{eq:hard_quantization}
\end{equation}
\begin{equation} C_{soft}( \tau_{a}, \tau_{b} ) = \sum_{i,j} exp(-\frac{d_{ij}^2}{\sigma^2}) Pr(i \in \tau_{a}) Pr( j\in \tau_{b} )
\label{eq:soft_quantization}
\end{equation}

Since $d_{i,j}$ decays exponentially, we compute equation~\ref{eq:soft_quantization} for $i,j$ that are at most $r$ pixels apart ( we use $r=3 \cdot \sigma$ ). In practice, this means that for each pixel, we evaluate $r^{2}$ pairs, and for each pair $k^{2}$ products of cluster assignment probabilities. This amounts to $O(n \cdot r^{2} \cdot k^{2} )$. In contrast, when evaluating equation~\ref{eq:hard_quantization} we have per pixel only $r^2$ non zeros pairs, which makes the complexity merely $O(n \cdot r^{2})$.

Normally, $Pr(i \in \tau_{a})$ is modeled as $K(p_i, \tau_{a})$ where $K$ is a kernel function (see Equation~\ref{eq:kernel}). We want a coarser model for $Pr(i \in \tau_{a})$ that will maintain the complexity of hard clustering. To do so, we assume that we have a hard clustering assignment $i \rightarrow \tau(i)$ and make the following approximation:

\begin{equation}
Pr( i \in \tau_{a} ) = K(p_i, \tau_{a}) \approx K( \tau(i), \tau_{a})
\label{eq:kernel}
\end{equation}

In words, the distance between pixel value $p_i$ and cluster $\tau_{a}$ is approximated by the distance between $\tau(i)$ and cluster $\tau_{a}$. Using this model we have:

\begin{align}
C_{soft}(\tau_{a},\tau_{b}) \\
= & \sum_{i,j} exp({-\frac{d_{ij}^2}{2\cdot\sigma^2}}) \cdot Pr( i \in \tau_{a} ) \cdot Pr( j \in \tau_{b} ) \notag \\ 
\underset{i}{\approx} & \sum_{i,j} exp({-\frac{d_{ij}^2}{2\cdot\sigma^2}}) \cdot K(\tau_{a},\tau(i)) \cdot K(\tau_{b},\tau(j)) \notag \\ 
 \underset{ii}{=} & \sum_{i,j} exp({-\frac{d_{ij}^2}{2\cdot\sigma^2}}) \cdot \sum_{\tau_{k_{1}}} [i \in \tau_{k_{1}}] \cdot \notag\\
 &K(\tau_{a},\tau_{k_{1}}) \cdot \sum_{\tau_{k_{2}}} [j \in \tau_{k_{2}}] \cdot K(\tau_{b},\tau_{k_{2}}) \notag \\
 \underset{iii}{=} &\sum_{\tau_{k_{1}},\tau_{k_{2}}} K(\tau_{a},\tau_{k_{1}}) \cdot K(\tau_{b},\tau_{k_{2}}) \cdot \notag\\
 & \sum_{i,j} exp({-\frac{d_{ij}^2}{2\cdot\sigma^2}}) \cdot [i \in \tau_{k_{1}}] \cdot [j \in \tau_{k_{2}}] \notag \\ 
 \underset{iv}{=} &\sum_{\tau_{k_{1}},\tau_{k_{2}}} K(\tau_{a},\tau_{k_{1}}) \cdot K(\tau_{b},\tau_{k_{2}}) \cdot C_{hard}(\tau_{k_{1}},\tau_{k_{2}}) \notag
\end{align}

where:
\begin{enumerate}[i]
	\item assign the approximation in equation (\ref{eq:kernel}). 
	\item $[i \in \tau_{k_{1}}]$ equals $1$ only for $\tau_{k_{1}} = \tau(i)$ and $0$ otherwise. 
	\item rearrange summations.
	\item use equation (\ref{eq:hard_quantization}) for hard quantization. 

\end{enumerate}

{\small
\bibliographystyle{ieee}
\bibliography{CoF}
}

\end{document}